\documentclass[10pt,twocolumn,letterpaper]{article}

\usepackage[pagenumbers]{cvpr} 

\usepackage{graphicx}
\usepackage{amsmath}
\usepackage{amssymb}
\usepackage{booktabs}
\usepackage{bbding}
\usepackage{pifont}
\usepackage{multirow}
\usepackage{makecell}
\usepackage[accsupp]{axessibility}  

\usepackage[pagebackref,breaklinks,colorlinks]{hyperref}

\usepackage[capitalize]{cleveref}
\crefname{section}{Sec.}{Secs.}
\Crefname{section}{Section}{Sections}
\Crefname{table}{Table}{Tables}
\crefname{table}{Tab.}{Tabs.}

\def\cvprPaperID{508} 
\def\confName{CVPR}
\def\confYear{2023}

\begin{document}

\title{Continuous Sign Language Recognition with Correlation Network}

\author{Lianyu Hu, Liqing Gao, Zekang Liu, Wei Feng\textsuperscript{\Envelope}
\and
College of Intelligence and Computing, Tianjin University, Tianjin 300350, China\\
Code : \url{https://github.com/hulianyuyy/CorrNet}
}
\maketitle

\begin{abstract}
Human body trajectories are a salient cue to identify actions in the video. Such body trajectories are mainly conveyed by hands and face across consecutive frames in sign language. However, current methods in continuous sign language recognition (CSLR) usually process frames independently, thus failing to capture cross-frame trajectories to effectively identify a sign. To handle this limitation, we propose correlation network (CorrNet) to explicitly capture and leverage body trajectories across frames to identify signs. In specific, a correlation module is first proposed to dynamically compute correlation maps between the current frame and adjacent frames to identify trajectories of all spatial patches. An identification module is then presented to dynamically emphasize the body trajectories within these correlation maps.  As a result, the generated features are able to gain an overview of local temporal movements to identify a sign. Thanks to its special attention on body trajectories, CorrNet achieves new state-of-the-art accuracy on four large-scale datasets, i.e., PHOENIX14, PHOENIX14-T, CSL-Daily, and CSL. A comprehensive comparison with previous spatial-temporal reasoning methods verifies the effectiveness of CorrNet. Visualizations demonstrate the effects of CorrNet on emphasizing human body trajectories across adjacent frames.
\end{abstract}

\section{Introduction}
Sign language is one of the most widely-used communication tools for the deaf community in their daily life. However, mastering this language is rather difficult and time-consuming for the hearing people, thus hindering direct communications between two groups. To relieve this problem, isolated sign language recognition tries to classify a video segment into an independent gloss\footnote{Gloss is the atomic lexical unit to annotate sign languages.}. Continuous sign language recognition (CSLR) progresses by sequentially translating images into a series of glosses to express a sentence, more prospective toward real-life deployment. 

Human body trajectories are a salient cue to identify actions in human-centric video understanding~\cite{wang2020video}. In sign language, such trajectories are mainly conveyed by both manual components (hand/arm gestures), and non-manual components (facial expressions, head movements, and body postures)~\cite{dreuw2007speech,ong2005automatic}. Especially, both hands move horizontally and vertically across consecutive frames quickly, with finger twisting and facial expressions to express a sign. To track and leverage such body trajectories is of great importance to understanding sign language. 

\begin{figure}[t]
  \centering
  \includegraphics[width=\linewidth]{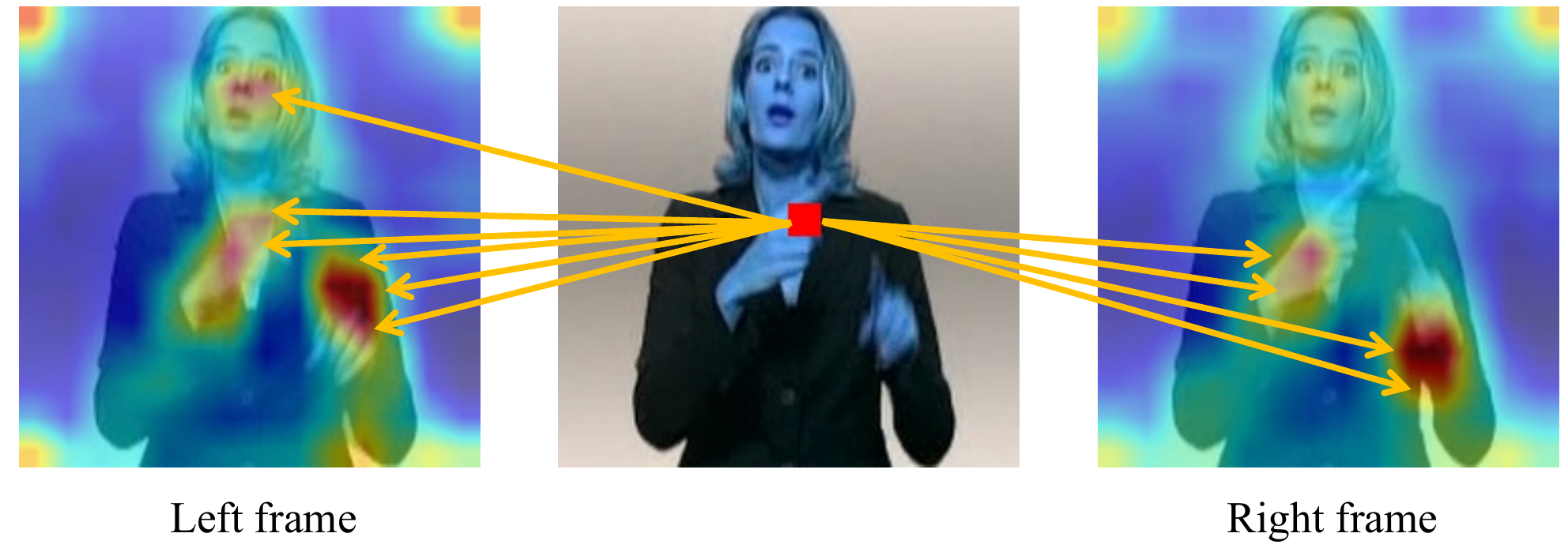}
  \caption{Visualization of correlation maps with Grad-CAM~\cite{selvaraju2017grad}. It's observed that without extra supervision, our method could well attend to informative regions in adjacent left/right frames to identify human body trajectories. }
  \label{fig0}
  \vspace{-7px}
\end{figure}

However, current CSLR methods~\cite{pu2020boosting,cheng2020fully,cui2019deep,niu2020stochastic,Min_2021_ICCV,zuo2022c2slr,hao2021self} usually process each frame separately, thus failing to exploit such critical cues in the early stage. Especially, they usually adopt a shared 2D CNN to capture spatial features for each frame independently. In this sense, frames are processed individually without interactions with adjacent neighbors, thus inhibited to identify and leverage cross-frame trajectories to express a sign. The generated features are thus not aware of local temporal patterns and fail to perceive the hand/face movements in expressing a sign. To handle this limitation, well-known 3D convolution~\cite{carreira2017quo} or its (2+1)D variants~\cite{tran2018closer,xie2018rethinking} are potential candidates to capture short-term temporal information to identify body trajectories. Other temporal methods like temporal shift~\cite{lin2019tsm} or temporal convolutions~\cite{liu2020teinet} can also attend to short-term temporal movements. However, it's hard for them to aggregate beneficial information from distant informative spatial regions due to their limited spatial-temporal receptive field. Besides, as their structures are fixed for each sample during inference, they may fail to dynamically deal with different samples to identify informative regions. To tackle these problems, we propose to explicitly compute correlation maps between adjacent frames to capture body trajectories, referred to as CorrNet. As shown in fig.~\ref{fig0}, our approach dynamically attends to informative regions in adjacent left/right frames to capture body trajectories, without relying on extra supervision. 

In specific, our CorrNet first employs a correlation module to compute correlation maps between the current frame and its adjacent frames to identify trajectories of all spatial patches. An identification module is then presented to dynamically identify and emphasize the body trajectories embodied within these correlation maps. This procedure doesn't rely on extra expensive supervision like body keypoints~\cite{zhou2020spatial} or heatmaps~\cite{zuo2022c2slr},  which could be end-to-end trained in a lightweight way. The resulting features are thus able to gain an overview of local temporal movements to identify a sign. Remarkably, CorrNet achieves new state-of-the-art accuracy on four large-scale datasets, i.e., PHOENIX14~\cite{koller2015continuous}, PHOENIX14-T~\cite{camgoz2018neural}, CSL-Daily~\cite{zhou2021improving}, and CSL~\cite{huang2018video}, thanks to its special attention on body trajectories. A comprehensive comparison with other spatial-temporal reasoning methods demonstrates the superiority of our method. Visualizations hopefully verify the effects of CorrNet on emphasizing human body trajectories across adjacent frames.

\section{Related Work}
\subsection{Continuous Sign Language Recognition}
Sign language recognition methods can be roughly categorized into isolated sign language recognition~\cite{tunga2021pose,hu2021signbert,hu2021hand} and continuous sign language recognition~\cite{pu2019iterative,cheng2020fully,cui2019deep,niu2020stochastic,Min_2021_ICCV} (CSLR), and we focus on the latter in this paper. CSLR tries to translate image frames into corresponding glosses in a weakly-supervised way: only sentence-level label is provided. Earlier methods~\cite{gao2004chinese,freeman1995orientation} in CSLR always employ hand-crafted features or HMM-based systems~\cite{koller2016deepsign,han2009modelling,koller2017re,koller2015continuous} to perform temporal modeling and translate sentences step by step. HMM-based systems first employ a feature extractor to capture visual features and then adopt an HMM to perform long-term temporal modeling. 

The recent success of convolutional neural networks (CNNs) and recurrent neural networks (RNNs) brings huge progress for CSLR.  The widely used CTC loss~\cite{graves2006connectionist} in recent CSLR methods~\cite{pu2019iterative,pu2020boosting,cheng2020fully,cui2019deep,niu2020stochastic,Min_2021_ICCV} enables training deep networks in an end-to-end manner by sequentially aligning target sentences with input frames. These CTC-based methods first rely on a feature extractor, i.e., 3D or 2D\&1D CNN hybrids, to extract frame-wise features, and then adopt a LSTM for capturing long-term temporal dependencies. However, several methods~\cite{pu2019iterative,cui2019deep} found in such conditions the feature extractor is not well-trained and then present an iterative training strategy to relieve this problem, but consume much more computations. Some recent studies~\cite{Min_2021_ICCV,cheng2020fully,hao2021self} try to directly enhance the feature extractor by adding alignment losses~\cite{Min_2021_ICCV,hao2021self} or adopt pseudo labels~\cite{cheng2020fully} in a lightweight way, alleviating the heavy computational burden. More recent works enhance CSLR by squeezing more representative temporal features~\cite{hu2022temporal} or dynamically emphasizing informative spatial regions~\cite{hu2023self}.

Our method is designed to explicitly incorporate body trajectories to identify a sign, especially those from hands and face. Some previous methods have also explicitly leveraged the hand and face features for better recognition. For example, CNN-LSTM-HMM~\cite{koller2019weakly} employs a multi-stream HMM (including hands and face) to integrate multiple visual inputs to improve recognition accuracy. STMC~\cite{zhou2020spatial} first utilizes a pose-estimation network to estimate human body keypoints and then sends cropped appearance regions (including hands and face) for information integration. More recently, C$^2$SLR~\cite{zuo2022c2slr} leverages the pre-extracted pose keypoints as supervision to guide the model to explicitly focus on hand and face regions. Our method doesn't rely on additional cues like pre-extracted body keypoints~\cite{zuo2022c2slr} or multiple streams~\cite{koller2019weakly}, which consume much more computations to leverage hand and face information. Instead, our model could be end-to-end trained to dynamically attend to body trajectories in a self-motivated way.

\subsection{Applications of Correlation Operation}
Correlation operation has been widely used in various domains, especially optical flow estimation and video action recognition. Rocco et al.~\cite{rocco2017convolutional} used it to estimate the geometric transformation between two images, and Feichtenhofer
et al.~\cite{feichtenhofer2017detect} applied it to capture object co-occurrences across time in tracking. For optical flow estimation, Deep matching~\cite{weinzaepfel2013deepflow} computes the correlation maps between image patches to find their dense correspondences. CNN-based methods like FlowNet~\cite{dosovitskiy2015flownet} and PWC-Net~\cite{sun2018pwc} design a correlation layer to help perform multiplicative patch comparisons between two feature maps. For video action recognition, Zhao et al.~\cite{zhao2018recognize} firstly employ a correlation layer to compute a cost volume to estimate the motion information. STCNet~\cite{diba2018spatio} considers spatial correlations and temporal correlations, respectively, inspired by SENet~\cite{hu2018squeeze}. MFNet~\cite{lee2018motion} explicitly estimates the approximation of optical flow based on fixed motion filters. Wang et al.~\cite{wang2020video} design a learnable correlation filter and replace 3D convolutions with the proposed filter to capture spatial-temporal information. Different from these methods that explicitly or implicitly estimate optical flow, the correlation operator in our method is used in combination with other operations to identify and track body trajectories across frames.

\section{Method}
\subsection{Overview}
\begin{figure}[t]
  \centering
  \includegraphics[width=\linewidth]{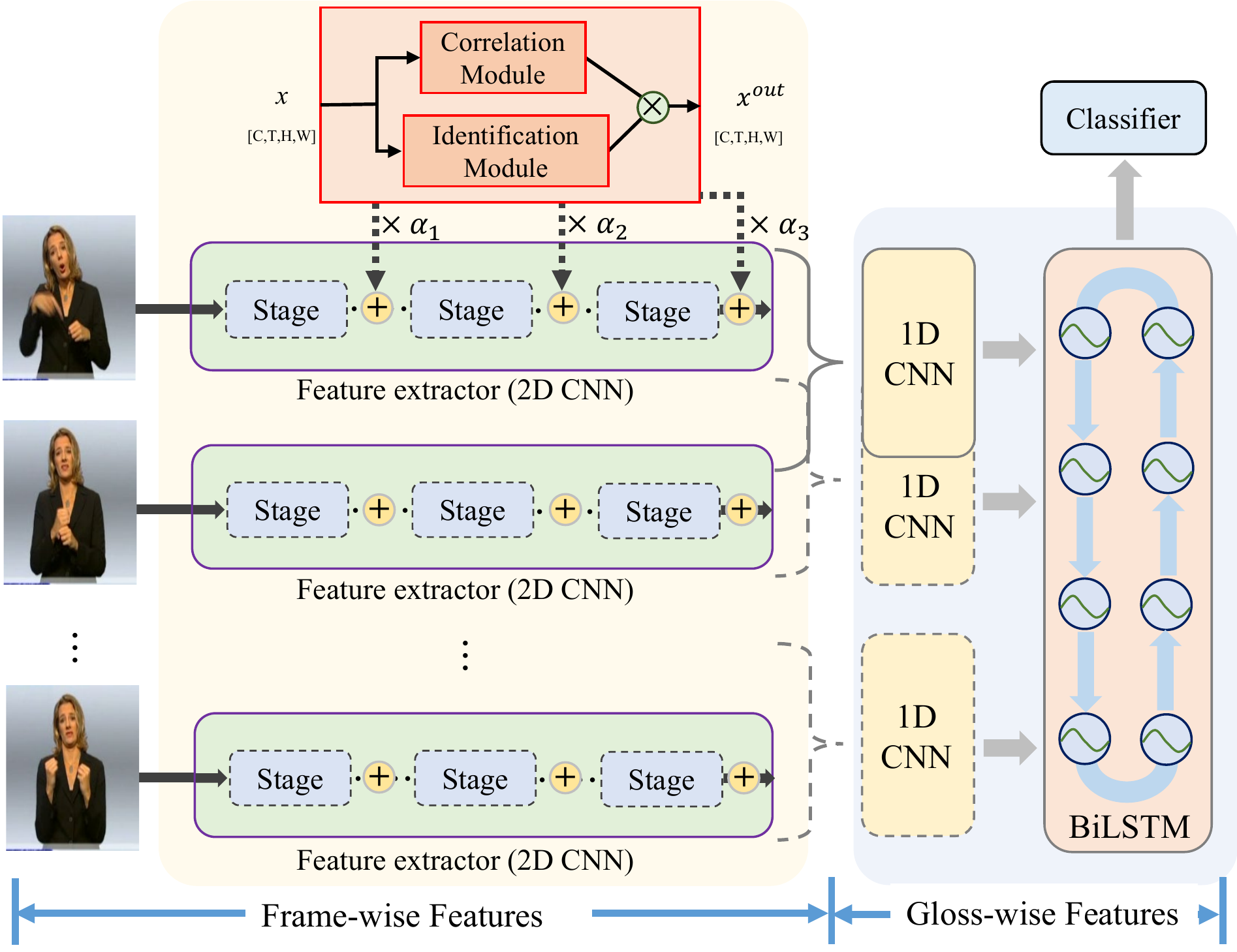}
  \caption{An overview for our CorrNet. It first employs a feature extractor (2D CNN) to capture frame-wise features, and then adopts a 1D CNN and a BiLSTM to perform short-term and long-term temporal modeling, respectively, followed by a classifier to predict sentences. We place our proposed identification module and correlation module after each stage of the feature extractor to identify body trajectories across adjacent frames.}
  \label{fig1}
  \end{figure}

As shown in fig.~\ref{fig1}, the backbone of CSLR models consists of a feature extractor (2D CNN\footnote{Here we only consider the feature extractor based on 2D CNN, because recent findings~\cite{adaloglou2021comprehensive,zuo2022c2slr} show 3D CNN can not provide as precise gloss boundaries as 2D CNN, and lead to lower accuracy. }), a 1D CNN, a BiLSTM, and a classifier (a fully connected layer) to perform prediction. Given a sign language video with $T$ input frames $x = \{x_{t}\}_{t=1}^T \in \mathcal{R}^{T \times 3\times H_0 \times W_0} $, a CSLR model aims to translate the input video into a series of glosses $y=\{ y_i\}_{i=1}^{N}$ to express a sentence, with $N$ denoting the length of the label sequence. Specifically, the feature extractor first processes input frames into frame-wise features $v = \{v_t\}_{t=1}^{T} \in \mathcal{R}^{T\times d}$. Then the 1D CNN and BiLSTM perform short-term and long-term temporal modeling based on these extracted visual representations, respectively. Finally, the classifier employs widely-used CTC loss~\cite{graves2006connectionist} to predict the probability of target gloss sequence $p(y|x)$.

The CSLR model processes input frames independently, failing to incorporate interactions between consecutive frames. We present a correlation module and an identification module to identify body trajectories across adjacent frames. Fig.~\ref{fig1} shows an example of a common feature extractor consisting of multiple stages. The proposed two modules are placed after each stage, whose outputs are element-wisely multiplied and added into the original features via a learnable coefficient $\alpha$. $\alpha$ controls the contributions of the proposed modules, and is initialized as zero to make the whole model keep its original behaviors. The correlation module computes correlation maps between consecutive frames to capture trajectories of all spatial patches. The identification module dynamically locates and emphasizes body trajectories embedded within these correlation maps. The outputs of correlation and identification modules are
multiplied to enhance inter-frame correlations.

\subsection{Correlation Module}
Sign language is mainly conveyed by both manual components (hand/arm gestures), and non-manual components (facial expressions, head movements, and body postures)~\cite{dreuw2007speech,ong2005automatic}. However, these informative body parts, e.g., hands or face, are misaligned in adjacent frames. We propose to compute correlation maps between adjacent frames to identify body trajectories. 

\begin{figure}[t]
  \centering
  \includegraphics[width=0.7\linewidth]{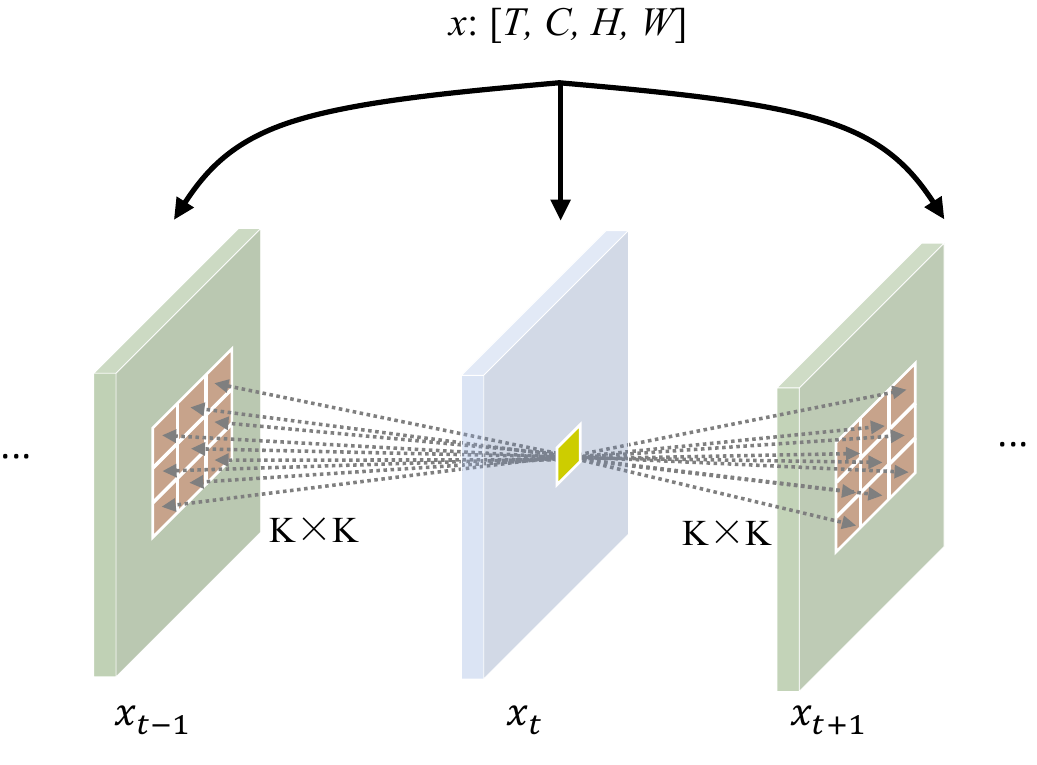}
  \caption{ Illustration for the correlation operator. It computes affinities between a feature patch $p(i,j)$ in $x_t$ and patches $p_{t+1}(i',j')$/$p_{t-1}(i',j')$ in adjacent frame $x_{t+1}$/$x_{t-1}$.}
  \label{fig2}
  \vspace{-5px}
  \end{figure}

Each frame could be represented as a 3D tensor $C\times H \times W$, where $C$ is the number of channels and $H\times W$ denotes spatial size. Given a feature patch $p_t(i,j)$ in current frame $x_t$, we compute the affinity between patch $p(i,j)$ and another patch $p_{t+1}(i',j')$ in adjacent frame $x_{t+1}$, where $(i,j)$ is the spatial location of the patch. To restrict the computation, the size of the feature patch could be reduced to a minimum, i.e., a pixel. The affinity between $p(i,j)$ and $p_{t+1}(i',j')$ is computed in a dot-product way as: 
\begin{equation}
  \label{e1}
  A(i,j,i',j') = \frac{1}{C} \sum_{c=1}^{C}{ (p^c_t(i,j) \cdot p^c_{t+1}(i',j')).}
\end{equation}
For the spatial location $(i,j)$ in $x_t$, $(i',j')$ is often restricted within a $K\times K$ neighborhood in $x_{t+1}$ to relieve spatial misalignment. A visualization is given in fig.~\ref{fig2}. Thus, for all pixels in $x_t$, the correlation maps are a tensor of size $H\times W\times K \times K$. $K$ could be set as a smaller value to keep semantic consistency or as a bigger value to attend to distant informative regions. 

Given the correlation map between a pixel and its neighbors in adjacent frame $x_{t+1}$, we constrain its range into (0,1) to measure their semantic similarity by passing $A(i,j,i',j')$ through a sigmoid function. We further subtract 0.5 from the results, to emphasize informative regions with positive values, and suppress redundant areas with negative values as:
\begin{equation}
  \label{e2}
  A'(i,j,i',j') = {\rm Sigmoid}(A(i,j,i',j'))-0.5
\end{equation}

After identifying the trajectories between adjacent frames, we incorporate these local temporal movements into the current frame $x_t$. Specifically, for a pixel in $x_t$, its trajectories are aggregated from its $K\times K$ neighbors in adjacent frame $x_{t+1}$, by multiplying their features with the corresponding affinities as :
\begin{equation}
  \label{e3}
   T(i,j) = \sum_{i',j'}{A'(i,j,i',j') * x_{t+1}(i',j').}
\end{equation}

In this sense, each pixel is able to be aware of its trajectories across consecutive frames. We aggregate bidirectional trajectories from both $x_{t-1}$ and $x_{t+1}$, and attach a learnable coefficient $\beta$ to measure the importance of bi-directions. Thus, eq.~\ref{e3} could be updated as :
\begin{equation}
  \begin{aligned}
  \label{e4}
   T(i,j) = & \beta_{1}\cdot \sum_{i',j'}{A_{t+1}'(i,j,i',j') * x_{t+1}(i',j')} + \\ & \beta_{2}\cdot \sum_{i',j'}{A_{t-1}'(i,j,i',j') * x_{t-1}(i',j')}
  \end{aligned}
\end{equation}
where $\beta_{1}$ and $\beta_{1}$ are initialized as 0.5.  This correlation calculation is repeated for each frame in a video to track body trajectories in videos.

\begin{figure}[t]
  \centering
  \includegraphics[width=\linewidth]{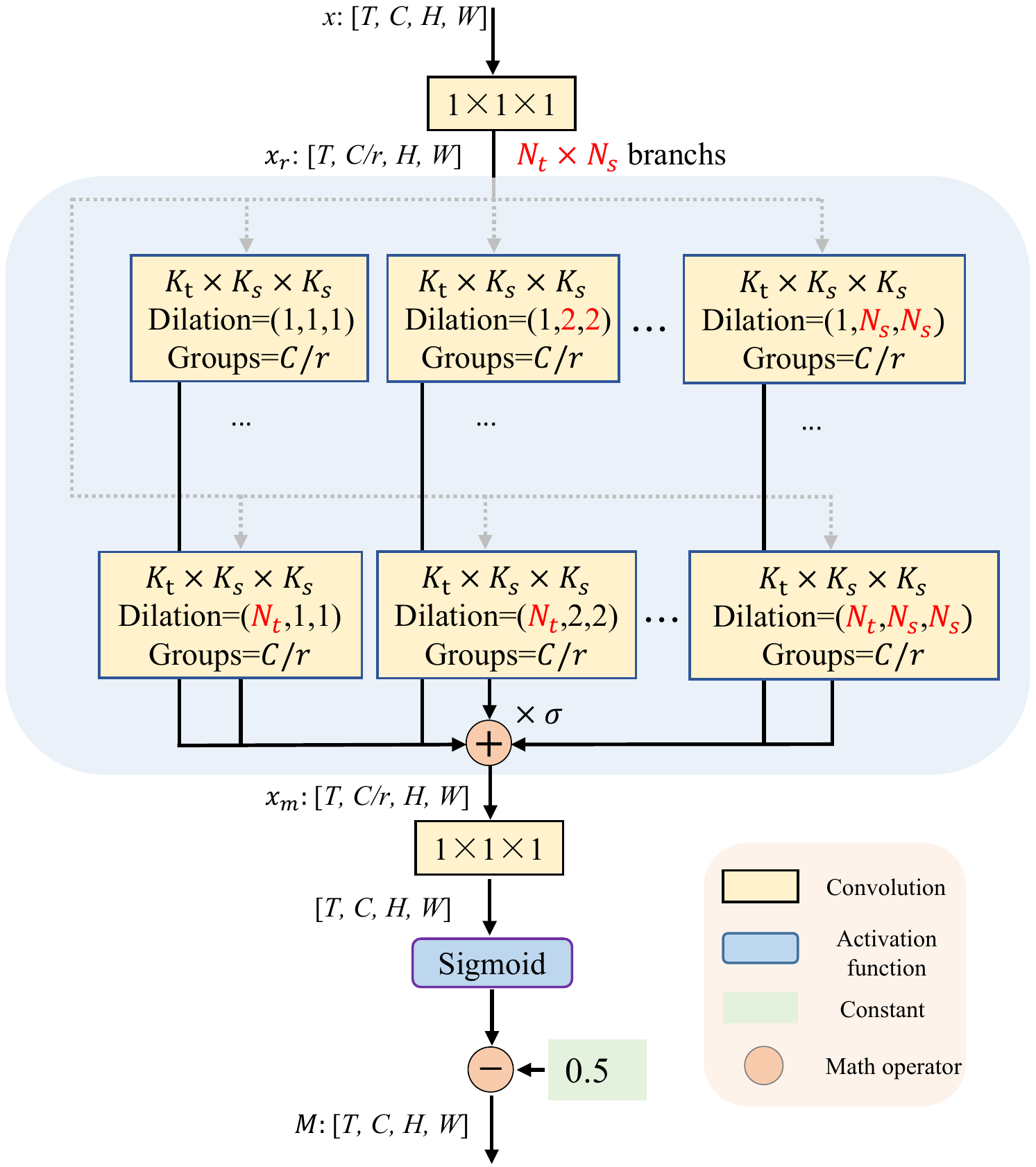}
  \caption{Illustration for our identification module.}
  \label{fig3}
  \end{figure}

\subsection{Identification Module}
The correlation module computes correlation maps between each pixel with its $K\times K$ neighbors in adjacent frames $x_{t-1}$ and $x_{t+1}$. However, as not all regions are critical for expressing a sign, only those informative regions carrying body trajectories should be emphasized in the current frame $x_t$. The trajectories of background or noise should be suppressed. We present an identification module to dynamically emphasize these informative spatial regions. Specifically, as informative regions like hand and face are misaligned in adjacent frames, the identification module leverages the closely correlated local spatial-temporal features to tackle the misalignment issue and locate informative regions. 

As shown in fig.~\ref{fig3}, the identification module first projects input features $x\in \mathcal{R}^{T \times C\times H \times W}$ into $x_r\in \mathcal{R}^{T \times C/r\times H \times W}$ with a $1\times 1\times 1$ convolution to decrease the computations, by a channel reduction factor $r$ as 16 by default.

As the informative regions, e.g., hands and face, are not exactly aligned in adjacent frames, it's necessary to consider a large spatial-temporal neighborhood to identify these features. Instead of directly employing a large 3D spatial-temporal kernel, we present a multi-scale paradigm by decomposing it into parallel branches of progressive dilation rates to reduce required computations and increase the model capacity. 

Specifically, as shown in fig.~\ref{fig3}, with a same small base convolution kernel of $K_t \times K_s \times K_s$, we employ multiple convolutions with their dilation rates increasing along spatial and temporal dimensions concurrently. The spatial and temporal dilation rate range within (1, $N_s$) and (1, $N_t$), respectively, resulting in total $N_s\times N_t$ branches. Group convolutions are employed for each branch to reduce parameters and computations. Features from different branches are multiplied with learnable coefficients \{$\sigma_1, \dots, \sigma_{N_s\times N_t}$\} to control their importance, and then added to mix information from branches of various spatial-temporal receptive fields as:
\begin{equation}
  \label{e6}
    x_m = \sum_{i=1}^{N_s}\sum_{j=1}^{N_t}{\sigma_{i,j} \cdot {\rm Conv}_{i,j}(x_r)}
\end{equation}
where the group-wise convolution ${\rm Conv}_{i,j}$ of different branches receives features of different spatial-temporal neighborhoods, with dilation rate $(j, i, i)$.

After receiving features from a large spatial-temporal neighborhood, $x_m$ is sent into a $1\times 1 \times 1$ convolution to project its channels back into $C$. It then passes through a sigmoid function to generate attention maps $M\in \mathcal{R}^{T \times C\times H \times W}$ with its values ranging within (0,1). Specially, $M$ is further subtracted from a constant value of 0.5 to emphasize informative regions with positive values, and suppress redundant areas with negative values as:
\begin{equation}
  \label{e7}
  M = {\rm Sigmoid}({\rm Conv}_{1\times 1\times 1}(x_m)) -0.5.
\end{equation}

Given the attention maps $M$ to identify informative regions, it's multiplied with the aggregated trajectories $T(x)$ by the correlation module to emphasize body trajectories and suppress others like background or noise. This refined trajectory information is finally incorporated into original spatial features $x$ via a residual connection as:
\begin{equation}
\label{e5}
 x^{out} = x + \alpha T(x) \cdot M.
\end{equation}
As stated before, $\alpha$ is initialized as zero to keep the original spatial features.


\section{Experiments}
\subsection{Experimental Setup}
\subsubsection{Datasets.} \textbf{PHOENIX14}~\cite{koller2015continuous} is recorded from a German weather forecast broadcast with nine actors before a clean background with a resolution of 210 $\times$ 260. It contains 6841 sentences with a vocabulary of 1295 signs, divided into 5672 training samples, 540 development (Dev) samples and 629 testing (Test) samples.

\textbf{PHOENIX14-T}~\cite{camgoz2018neural} is available for both CSLR and sign language translation tasks. It contains 8247 sentences with a vocabulary of 1085 signs, split into 7096 training instances, 519 development (Dev) instances and 642 testing (Test) instances.

\textbf{CSL-Daily}~\cite{zhou2021improving} revolves the daily life, recorded indoor at 30fps by 10 signers. It contains 20654 sentences, divided into 18401 training samples, 1077 development (Dev) samples and 1176 testing (Test) samples. 

\textbf{CSL}~\cite{huang2018video} is collected in the laboratory environment by fifty signers with a vocabulary size of 178 with 100 sentences. It contains 25000 videos, divided into training and testing sets by a ratio of 8:2.

\subsubsection{Training details.} For fair comparisons, we follow the same setting as state-of-the-art methods~\cite{Min_2021_ICCV,zuo2022c2slr} to prepare our model. We adopt ResNet18~\cite{he2016deep} as the 2D CNN backbone with ImageNet~\cite{deng2009imagenet} pretrained weights. The 1D CNN of state-of-the-art methods is set as a sequence of \{K5, P2, K5, P2\} layers where K$\sigma$ and P$\sigma$ denotes a 1D convolutional layer and a pooling layer with kernel size of $\sigma$, respectively. A two-layer BiLSTM with hidden size 1024 is attached for long-term temporal modeling, followed by a fully connected layer for sentence prediction. We train our models for 40 epochs with initial learning rate 0.001 which is divided by 5 at epoch 20 and 30. Adam~\cite{kingma2014adam} optimizer is adopted as default with weight decay 0.001 and batch size 2. All input frames are first resized to 256$\times$256, and then randomly cropped to 224$\times$224 with 50\% horizontal flipping and 20\% temporal rescaling during training. During inference, a 224$\times$224 center crop is simply adopted. Following VAC~\cite{Min_2021_ICCV}, we employ the VE loss and VA loss for visual supervision, with weights 1.0 and 25.0, respectively. Our model is trained and evaluated upon a 3090 graphical card.

\subsubsection{Evaluation Metric.} We use Word Error Rate (WER) as the evaluation metric, which is defined as the minimal summation of the \textbf{sub}stitution, \textbf{ins}ertion, and \textbf{del}etion operations to convert the predicted sentence to the reference sentence, as:
\begin{equation}
\label{e11}
\rm WER = \frac{ \#sub+\#ins+\#del}{\#reference}.
\end{equation}
Note that the \textbf{lower} WER, the \textbf{better} accuracy.

\subsection{Ablation Study}
We report ablative results on both development (Dev) and testing (Test) sets of PHOENIX14 dataset.

\begin{table}[t]   
  \centering
  \begin{tabular}{cccc}
  \hline
  Configurations & Dev(\%) & Test(\%)\\
  \hline
  - & 20.2 & 21.0\\
  \hline
  $N_t$=4, $N_s$=\textbf{1}  & 19.6  & 20.1\\
  $N_t$=4, $N_s$=\textbf{2} & 19.2 & 19.8 \\
  $N_t$=4, $N_s$=\textbf{3} & \textbf{18.8} & \textbf{19.4} \\
  $N_t$=4, $N_s$=\textbf{4}  & 19.1  &19.7 \\
  \hline
  $N_t$=\textbf{2}, $N_s$=3 & 19.4  & 19.9 \\
  $N_t$=\textbf{3}, $N_s$=3 & 19.1  & 19.7 \\
  $N_t$=\textbf{4}, $N_s$=3 & \textbf{18.8} & \textbf{19.4} \\
  $N_t$=\textbf{5}, $N_s$=3  & 19.3 & 19.8 \\
  \hline
  $K_t$=\textbf{9}, $K_s$=\textbf{7} & 19.9 & 20.4 \\
  \hline
  \end{tabular}
  \caption{Ablations for the multi-scale architecture of identification module on the PHOENIX14 dataset.} 
  \label{tab1} 
  \end{table}

\textbf{Study on the multi-scale architecture of identification module.} In tab.~\ref{tab1}, without identification module, our baseline achieves 20.2\% and 21.0\% WER on the Dev and Test Set, respectively. The base kernel size is set as $3\times 3\times 3$ for $K_t\times K_s\times K_s$. When fixing $N_t$=4 and varying spatial dilation rates to expand spatial receptive fields, it's observed a larger $N_s$ consistently brings better accuracy. When $N_s$ reaches 3, it brings no more accuracy gain. We set $N_s$ as 3 by default and test the effects of $N_t$. One can see that either increasing $K_t$ to 5 or decreasing $K_t$ to 2 and 3 achieves worse accuracy. We thus adopt $N_t$ as 4 by default. We also compare our proposed multi-scale architecture with a normal implementation of more parameters. The receptive field of the identification module with $N_t$=4, $N_s$=3 is identical to a normal convolution with $K_t$=9 and $K_s$=7. As shown in the bottom of tab.~\ref{tab1}, although a normal convolution owns more parameters and computations than our proposed architecture, it still performs worse, verifying the effectiveness of our architecture.

\begin{table}[t]   
  \centering
  \begin{tabular}{ccc}
  \hline
  Configurations  & Dev(\%) & Test(\%)\\
  \hline
  - & 20.2 & 21.0\\
  $K$=\textbf{3} & 19.6  & 20.4 \\
  $K$=\textbf{5} & 19.4 & 20.2 \\
  $K$=\textbf{7} & 19.2 & 20.0 \\
  $K$=\textbf{9} & 19.1  & 19.8 \\
  $K$= $H$ or $W$ (Full image) & \textbf{18.8} & \textbf{19.4} \\
  \hline
  \end{tabular}
  \caption{Ablations for the articulated area of correlation module on the PHOENIX14 dataset.} 
  \label{tab2} 
  \end{table}

\textbf{Study on the neighborhood $K$ of correlation module.} In tab.~\ref{tab2}, when $K$ is null, the correlation module is disabled. It's observed that a larger $K$, i.e., more incorporated spatial-temporal neighbors, consistently brings better accuracy. The performance reaches the peak when $K$ equals $H$ or $W$, i.e., the full image is incorporated. In this case, distant informative objects could be interacted to provide discriminative information. We set $K$= $H$ or $W$ by default.

\begin{table}[t]   
  \centering
  \begin{tabular}{cccc}
  \hline
  \hline
  Correlation & Identification & Dev(\%) & Test(\%)\\
  \hline
  \ding{56} & \ding{56}   & 20.2 & 21.0 \\
  \Checkmark & \ding{56}     & 19.5 & 20.0 \\
  \ding{56} & \Checkmark    & 19.4 & 19.9\\
  \Checkmark & \Checkmark    & \textbf{18.8} & \textbf{19.4} \\
  \hline
  \end{tabular}
  \caption{Ablations for the effectiveness of correlation module and identification module on the PHOENIX14 dataset.} 
  \label{tab3} 
  \end{table}

\textbf{Effectiveness of two proposed modules.} In tab.~\ref{tab3}, we first notice that either only using the correlation module or identification module could already bring a notable accuracy boost, with 19.5\% \& 20.0\% and 19.4\% \& 19.9\% accuracy on the Dev and Test Sets, respectively. When combining both modules, the effectiveness is further activated with 18.8\% \& 19.4\% accuracy on the Dev and Test Sets, respectively, which is adopted as the default setting.
  
\begin{table}[t]   
  \centering
  \begin{tabular}{ccccc}
  \hline
  Stage 2 & Stage 3 & Stage 4 & Dev(\%) & Test(\%)\\
  \hline
  \ding{56} & \ding{56}  & \ding{56}    & 20.2 & 21.0 \\
    \Checkmark & \ding{56}  & \ding{56}   & 19.6 & 20.1 \\
    \ding{56} & \Checkmark  & \ding{56}   & 19.5 & 20.2\\
    \ding{56} & \ding{56}  & \Checkmark   & 19.4 & 20.0 \\
    \hline
    \Checkmark & \Checkmark  & \ding{56}   & 19.2 & 19.9 \\
    \Checkmark & \Checkmark  & \Checkmark   & \textbf{18.8} & \textbf{19.4} \\
  \hline
  \end{tabular}
  \caption{Ablations for the locations of CorrNet on the PHOENIX14 dataset.} 
  \label{tab4} 
  \end{table}

\textbf{Effects of locations for CorrNet.} Tab~\ref{tab4} ablates the locations of our proposed modules, which are placed after Stage 2, 3 or 4. It's observed that choosing any one of these locations could bring a notable accuracy boost, with 19.6\% \& 20.1\%, 19.5\% \& 20.2\% and 19.4\% \& 20.0\% accuracy boost. When combining two or more locations, a larger accuracy gain is witnessed. The accuracy reaches the peak when proposed modules are placed after Stage 2, 3 and 4, with 18.8\% \& 19.4\% accuracy, which is adopted by default.

\begin{table}[t]   
  \centering
  \begin{tabular}{lcc}
  \hline
  Configurations & Dev(\%) & Test(\%)\\
  \hline
  SqueezeNet~\cite{hu2018squeeze}  & 22.2 & 22.6 \\
  \quad w/ CorrNet  & \textbf{20.2}  & \textbf{20.4} \\
  \hline
  ShuffleNet V2~\cite{ma2018shufflenet} & 21.7 & 22.2 \\
  \quad w/ CorrNet  & \textbf{19.7}  & \textbf{20.2} \\
  \hline
  GoogleNet~\cite{szegedy2015going}  & 21.4 & 21.5\\
  \quad w/ CorrNet  & \textbf{19.6} & \textbf{19.8} \\
  \hline
  \end{tabular}
  \caption{Ablations for the generalizability of CorrNet over multiple backbones on the PHOENIX14 dataset.} 
  \label{tab5} 
  \end{table}

\textbf{Generalizability of CorrNet.} We deploy CorrNet upon multiple backbones, including SqueezeNet~\cite{hu2018squeeze}, ShuffleNet V2~\cite{ma2018shufflenet} and GoogLeNet~\cite{szegedy2015going} to validate its generalizability in tab.~\ref{tab5}. The proposed modules are placed after three spatial downsampling layers in SqueezeNet, ShuffleNet V2 and GoogLeNet, respectively. It's observed that our proposed model generalizes well upon different backbones, bringing +2.0\% \& +2.2\%, +2.0\% \& +2.0\% and +1.8\% \& +1.7\% accuracy boost on the Dev and Test Sets, respectively.

  \begin{table}[t]   
  \centering
  \setlength\tabcolsep{3pt}
  \begin{tabular}{lcc}
  \hline
  Methods & Dev(\%) & Test(\%)\\
  \hline
  - & 20.2 & 21.0\\
  w/ SENet~\cite{hu2018squeeze}  & 19.8  & 20.4 \\
  w/ CBAM~\cite{woo2018cbam} & 19.7 & 20.2 \\
  w/ NLNet~\cite{wang2018non} & - & -\\
  \hline
  I3D~\cite{carreira2017quo} & 22.6  & 22.9 \\
  R(2+1)D~\cite{tran2018closer}  & 22.4  & 22.3 \\
  TSM~\cite{lin2019tsm} & 19.9 & 20.5 \\
  \hline
  CorrNet & \textbf{18.8} & \textbf{19.4} \\
  \hline
  \end{tabular}
  \caption{Comparison with other methods of spatial-temporal attention or temporal reasoning on the PHOENIX14 dataset.} 
  \label{tab6} 
  \end{table}

\begin{table}[t]   
  \centering
  \setlength\tabcolsep{3pt}
  \begin{tabular}{lcc}
  \hline
  Methods & Dev(\%) & Test(\%)\\
  \hline
  CNN+HMM+LSTM~\cite{koller2019weakly} & 26.0  & 26.0 \\
  DNF~\cite{cui2019deep} & 23.1 & 22.9\\
  STMC~\cite{zhou2020spatial} & 21.1  & 20.7 \\
  C$^2$SLR~\cite{zuo2022c2slr} & 20.5  & 20.4 \\
  \hline
  CorrNet & \textbf{18.8} & \textbf{19.4} \\
  \hline
  \end{tabular}
  \caption{Comparison with other methods that explicitly exploit hand and face features on the PHOENIX14 dataset.} 
  \label{tab7} 
  \end{table}

\begin{figure*}[t]
  \centering
  \includegraphics[width=\linewidth]{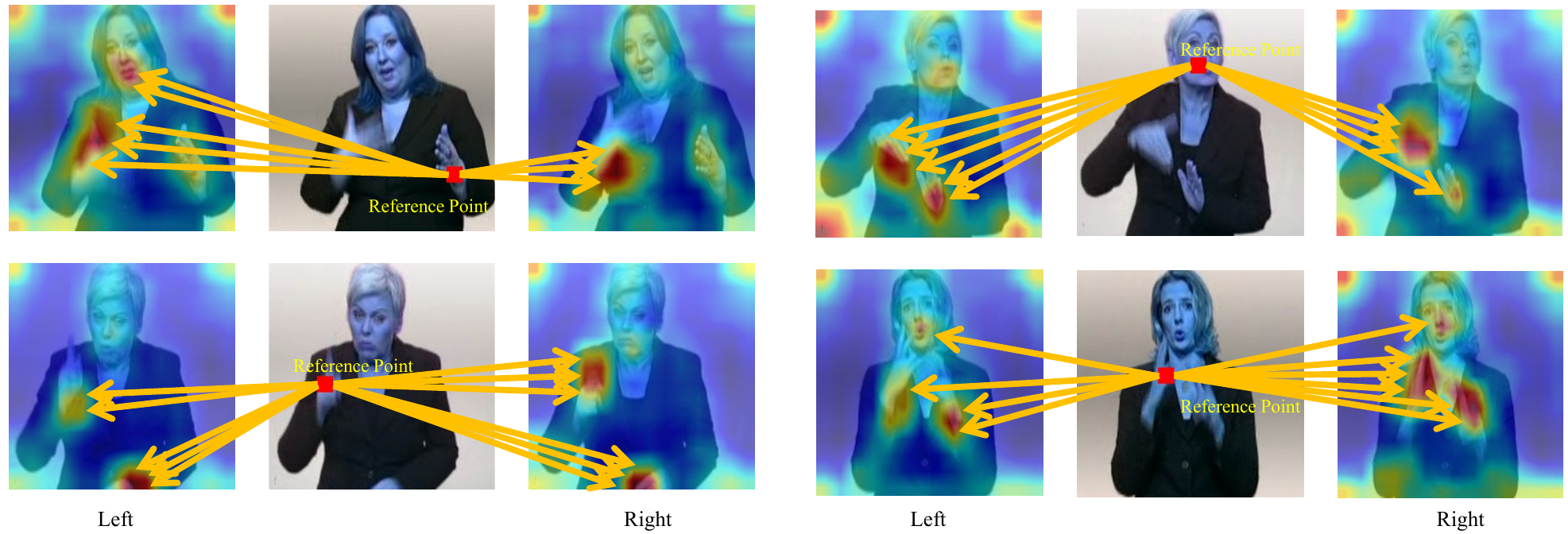}
  \caption{Visualizations of correlation maps for correlation module. Based on correlation operators, each frame could especially attend to informative regions in adjacent left/right frames like hands and face (dark red areas).}
  \label{fig4}
  \end{figure*}  

\textbf{Comparisons with other spatial-temporal reasoning methods.} 
Tab.~\ref{tab6} compares our approach with other methods of spatial-temporal reasoning ability. SENet~\cite{hu2018squeeze} and CBAM~\cite{woo2018cbam} perform channel attention to emphasize key information. NLNet~\cite{wang2018non} employs non-local means to aggregate spatial-temporal information from other frames. I3D~\cite{carreira2017quo} and R(2+1)D~\cite{tran2018closer} deploys 3D or 2D+1D convolutions to capture spatial-temporal features. TSM~\cite{lin2019tsm} adopts temporal shift operation to obtain features from adjacent frames. In the upper part of tab.~\ref{tab6}, one can see CorrNet largely outperforms other attention-based methods, i.e., SENet, CBAM and NLNet, for its superior ability to identify and aggregate body trajectories. NLNet is out of memory due to its quadratic computational complexity with spatial-temporal size. In the bottom part of tab.~\ref{tab6}, it's observed that I3D and R(2+1)D even degrade accuracy, which may be attributed to their limited spatial-temporal receptive fields and increased training complexity. TSM slightly brings 0.3\% \& 0.3\% accuracy boost. Our proposed approach surpasses these methods greatly, verifying its effectiveness in aggregating beneficial spatial-temporal information, from even distant spatial neighbors. 

\begin{figure}[t]
  \centering
  \includegraphics[width=\linewidth]{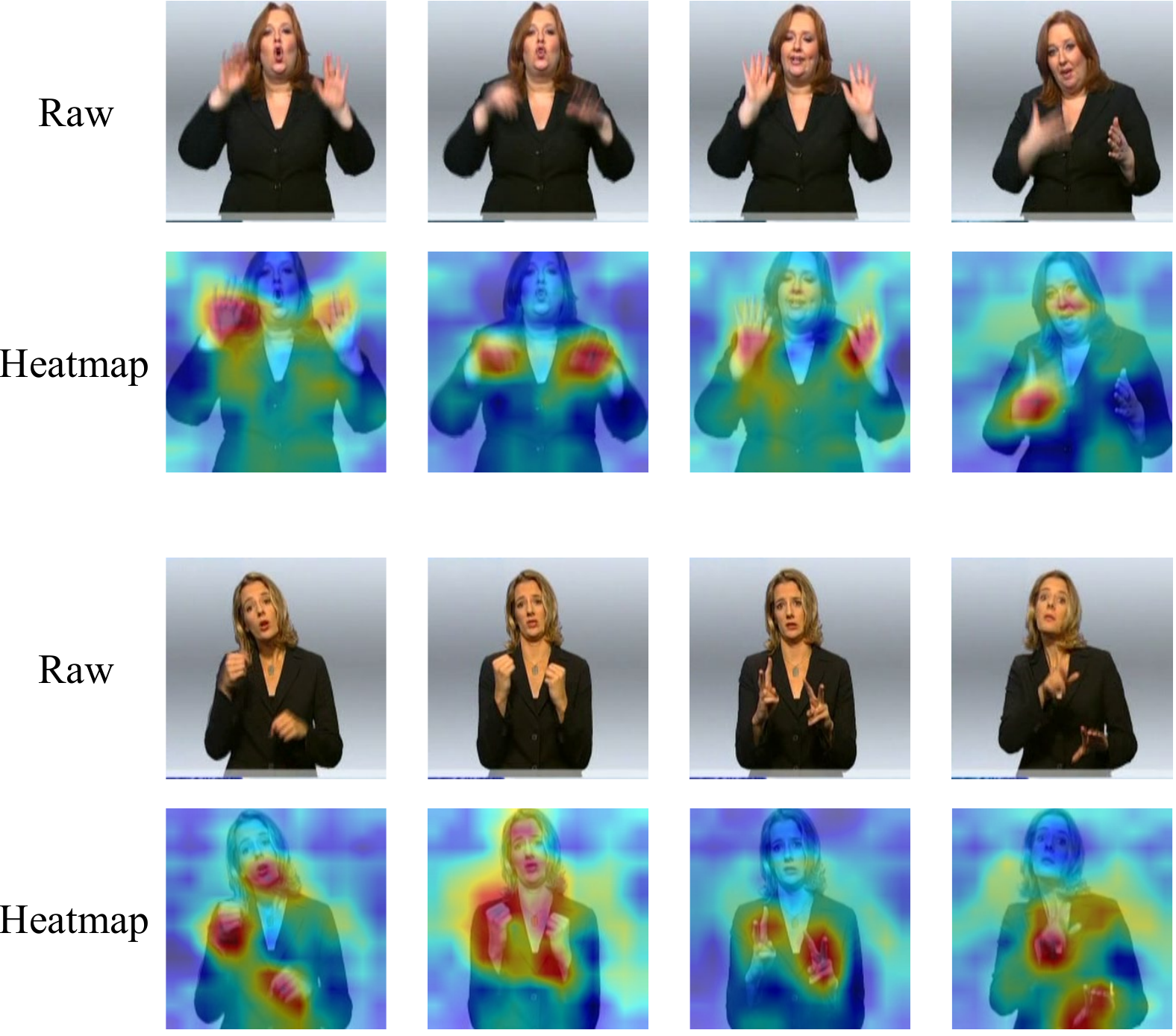}
  \caption{Visualizations of heatmaps by Grad-CAM~\cite{selvaraju2017grad}. Top: raw frames; Bottom: heatmaps of our identification module. Our identification module could generally focus on the human body (light yellow areas) and especially pays attention to informative regions like hands and face (dark red areas) to track body trajectories.}
  \label{fig5}
  \end{figure} 

\textbf{Comparisons with previous methods equipped with hand or face features.} Many previous CSLR methods explicitly leverage hand and face features for better recognition, like multiple input streams~\cite{koller2019weakly}, human body keypoints~\cite{zhou2020spatial,zuo2022c2slr} and pre-extracted hand patches~\cite{cui2019deep}. They require extra expensive pose-estimation networks like HRNet~\cite{wang2020deep} or additional training stages. Our approach doesn't rely on extra supervision and could be end-to-end trained to dynamically attend to body trajectories like hand and face in a self-motivated way. Tab.~\ref{tab7} shows that our method outperforms these methods by a large margin.

\begin{table*}[t]   
  \centering
  \setlength\tabcolsep{3pt}
  \begin{tabular}{ccccccccc}
  \hline
  \multirow{3}{*}{Methods} &\multirow{3}{*}{Backbone} & \multicolumn{4}{c}{PHOENIX14} & \multicolumn{2}{c}{PHOENIX14-T} \\
  & &\multicolumn{2}{c}{Dev(\%)} & \multicolumn{2}{c}{Test(\%)} &  \multirow{2}{*}{Dev(\%)} & \multirow{2}{*}{Test(\%)}\\
  & &del/ins & WER & del/ins& WER & & \\
  \hline
  SFL~\cite{niu2020stochastic}& ResNet18 & 7.9/6.5 & 26.2 & 7.5/6.3& 26.8 & 25.1&26.1\\
  FCN~\cite{cheng2020fully}& Custom & - & 23.7 & -& 23.9 & 23.3& 25.1\\
  CMA~\cite{pu2020boosting} & GoogLeNet & 7.3/2.7 & 21.3 & 7.3/2.4 & 21.9  & -&-\\
  VAC~\cite{Min_2021_ICCV}& ResNet18 & 7.9/2.5 & 21.2 &8.4/2.6 & 22.3 &- &-\\
  SMKD~\cite{hao2021self}& ResNet18 &6.8/2.5 &20.8 &6.3/2.3 & 21.0 & 20.8 & 22.4\\
  TLP~\cite{hu2022temporal} & ResNet18 & 6.3/2.8 & 19.7 & 6.1/2.9 & 20.8 & 19.4  & 21.2 \\
  SEN~\cite{hu2023self} & ResNet18 & 5.8/2.6 &  19.5 &  7.3/4.0 &  21.0 &  19.3 &  20.7 \\
  \hline
  SLT$^*$~\cite{camgoz2018neural}& GoogLeNet  & - & - & - & - & 24.5 & 24.6\\
  CNN+LSTM+HMM$^*$~\cite{koller2019weakly}& GoogLeNet  & - &26.0 & - & 26.0 & 22.1 & 24.1 \\
  DNF$^*$~\cite{cui2019deep}& GoogLeNet  & 7.3/3.3 &23.1& 6.7/3.3 & 22.9 & - & -\\
  STMC$^*$~\cite{zhou2020spatial}& VGG11 & 7.7/3.4 &21.1 & 7.4/2.6 & 20.7 & 19.6 & 21.0\\
  C$^2$SLR$^*$~\cite{zuo2022c2slr} & ResNet18 & - & 20.5 &- & 20.4 & 20.2 & 20.4  \\
  \hline
  \textbf{CorrNet } & ResNet18 & 5.6/2.8  &\textbf{18.8} &   5.7/2.3 & \textbf{19.4}  & \textbf{18.9} & \textbf{20.5} \\
  \hline   
  \end{tabular}  
  \caption{Comparison with state-of-the-art methods on the PHOENIX14 and PHOENIX14-T datasets. $*$ indicates extra clues such as face or hand features are included by additional networks or pre-extracted heatmaps.} 
  \label{tab8}
\end{table*}

\subsection{Visualizations}
\textbf{Visualizations for correlation module.} Fig.~\ref{fig4} shows the correlation maps generated by our correlation module with adjacent frames. It's observed that the reference point could well attend to informative regions in adjacent left/right frame, e.g., hands or face, to track body trajectories in expressing a sign. Especially, they always focus on the moving body parts that play a major role in expressing signs. For example, the reference point (left hand) in the upper left figure specially attends to the quickly moving right hand to capture sign information.

\textbf{Visualizations for identification module.} Fig.~\ref{fig5} shows the heatmaps generated by our identification module. Our identification module could generally focus on the human body (light yellow areas). Especially, it pays major attention to regions like hands and face (dark red areas). These results show that our identification module could dynamically emphasize important areas in expressing a sign, e.g., hands and face, and suppress other regions. 

\subsection{Comparison with State-of-the-Art Methods}
\textbf{PHOENIX14} and \textbf{PHOENIX14-T}. Tab.~\ref{tab8} shows a comprehensive comparison between our CorrNet and other state-of-the-art methods. The entries notated with $*$ indicate these methods utilize additional factors like face or hand features for better accuracy. We notice that CorrNet outperforms other state-of-the-art methods by a large margin upon both datasets, thanks to its special attention on body trajectories. Especially, CorrNet outperforms previous CSLR methods equipped with hand and faces acquired by heavy pose-estimation networks or pre-extracted heatmaps (notated with *), without additional expensive supervision. 

\textbf{CSL-Daily}. CSL-Daily is a recently released large-scale dataset with the largest vocabulary size (2k) among commonly-used CSLR datasets, with a wide content covering family life, social contact and so on. Tab.~\ref{tab9} shows that our CorrNet achieves new state-of-the-art accuracy upon this challenging dataset with notable progress, which generalizes well upon real-world scenarios.

\textbf{CSL}. As shown in tab.~\ref{tab10}, our CorrNet could achieve extremely superior accuracy (0.8\% WER) upon this well-examined dataset, outperforming existing CSLR methods.

\begin{table}[t]   
  \centering
  \setlength\tabcolsep{2pt}
  \begin{tabular}{cccc}
  \hline
  Methods&  Dev(\%) & Test(\%)\\
  \hline
  LS-HAN~\cite{huang2018video}  & 39.0  & 39.4\\
  TIN-Iterative~\cite{cui2019deep}  & 32.8  & 32.4\\
  Joint-SLRT~\cite{camgoz2020sign}  & 33.1  & 32.0 \\
  FCN~\cite{cheng2020fully} & 33.2  & 32.5 \\
  BN-TIN~\cite{zhou2021improving} & 33.6  & 33.1 \\
  \hline
  \textbf{CorrNet} & \textbf{30.6} & \textbf{30.1} \\
  \hline
  \end{tabular}  
  \caption{Comparison with state-of-the-art methods on the CSL-Daily dataset~\cite{zhou2021improving}.} 
  \label{tab9}
  \end{table}

\begin{table}[t]   
  \centering
  \setlength\tabcolsep{2pt}
  \begin{tabular}{cc}
    \hline
    Methods&  WER(\%)\\
    \hline
    LS-HAN~\cite{huang2018video}  & 17.3 \\
    SubUNet~\cite{cihan2017subunets}   & 11.0\\
    SF-Net~\cite{yang2019sf} & 3.8 \\
    FCN~\cite{cheng2020fully}   & 3.0 \\
    STMC~\cite{zhou2020spatial}  & 2.1 \\
    VAC~\cite{Min_2021_ICCV} & 1.6 \\
    C$^2$SLR~\cite{zuo2022c2slr} & 0.9 \\
    \hline
    \textbf{CorrNet} & \textbf{0.8} \\
    \hline
    \end{tabular}  
    \caption{Comparison with state-of-the-art methods on the CSL dataset~\cite{huang2018video}.} 
    \label{tab10}
    \vspace{-5px}
  \end{table}

\section{Conclusion}
This paper introduces a correlation module to capture trajectories between adjacent frames and an identification module to locate body regions. Comparisons with previous CSLR methods with spatial-temporal reasoning ability or equipped with hand and face features demonstrate the superiority of CorrNet. Visualizations show that CorrNet could generally attend to hand and face regions to capture body trajectories.
{\small
\bibliographystyle{ieee_fullname}
\bibliography{ref}
}

\end{document}